\documentclass[sigconf]{acmart}

\usepackage{booktabs} 

\setcopyright{rightsretained}
\usepackage{subfigure}
\usepackage{algorithm}
\usepackage{algorithmic}

\acmDOI{10.475/123_4}

\acmISBN{123-4567-24-567/08/06}

\acmConference[MM'17]{ACM MM}{October 23-27 2017}{ Mountain View, CA USA}

\acmPrice{15.00}

\author{Hantao Yao$^{1,2}$, Shiliang Zhang$^{3}$, Yongdong Zhang$^{1,2}$, Jintao Li$^{1}$, Qi Tian$^{4}$}
\affiliation{%
\small
  \institution{
$^{1}$Key Lab of Intelligent Information Processing of CAS, Institute of Computing Technology, CAS, Beijing 100190, China \\
 $^{2}$ University of Chinese Academy of Sciences, Beijing 100049, China \\
 $^{3}$School of Electronic Engineering and Computer Science, Peking University, Beijing 100871, China \\
  $^{4}$ Department of Computer Science University of Texas at San Antonio, San Antonio, USA \\}
}
\email{{yaohantao,zhyd,jtli}@ict.ac.cn, slzhang.jdl@pku.edu.cn, qitian@cs.utsa.edu}

\begin{document}

\title{One-Shot Fine-Grained Instance Retrieval}

\begin{abstract}
Fine-Grained Visual Categorization (FGVC) has achieved significant progress recently. However, the number of fine-grained species could be huge and dynamically increasing in real scenarios, making it difficult to recognize unseen objects under the current FGVC framework. This raises an open issue to perform large-scale fine-grained identification without a complete training set. Aiming to conquer this issue, we propose a retrieval task named One-Shot Fine-Grained Instance Retrieval (OSFGIR). ``One-Shot'' denotes the ability of identifying unseen objects through a fine-grained retrieval task assisted with an incomplete auxiliary training set. This paper first presents the detailed description to OSFGIR task and our collected \emph{OSFGIR-378K} dataset. Next, we propose the Convolutional and Normalization Networks (CN-Nets) learned on the auxiliary dataset to generate a concise and discriminative representation. Finally, we present a coarse-to-fine retrieval framework consisting of three components, \emph{i.e.}, coarse retrieval, fine-grained retrieval, and query expansion, respectively. The framework progressively retrieves images with similar semantics, and performs fine-grained identification. Experiments show our OSFGIR framework achieves significantly better accuracy and efficiency than existing FGVC and image retrieval methods, thus could be a better solution for large-scale fine-grained object identification. 
\end{abstract}

%
%



\keywords{One-Shot Fine-Grained Instance Retrieval, Fine-Grained Visual Categorization, CNN, CN-Nets, OSFGIR-378K}

\maketitle

\section{Introduction}

Different from conventional object categorization, Fine-Grained Visual Categorization (FGVC) aims to identify objects belonging to the same or closely-related species that only experienced experts can recognize, \emph{e.g.}, identify a bird as ``Black footed Albatross'' or ``Sooty Albatross''. Due to the ability of providing valuable information to users, FGVC has been attracting lots of attentions~\cite{berg2013poof,branson2014bird,chai2013symbiotic, ge2015subset,gavves2014local, gavves2013fine,liu2013bird, liu2014part, xiao2014application, zhang2014part,simon2015neural,lin2015deep,krause2015fine,liu2016localizing,lin2015bilinear}. Although FGVC is challenging, its performance has been significantly improved by using powerful Convolutional Neural Networks~\cite{zhang2014part,lin2015bilinear,krause2015fine,jaderberg2015spatial}, considering detailed part localization~\cite{lin2015deep,zhang2014part,simon2015neural,liu2016localizing,branson2014bird}, and generating better visual descriptions~\cite{berg2013poof,simon2015neural,krause2015fine,liu2016localizing}. For instance, the classification accuracy on CUB-200-2011~\cite{wah2011caltech} has been pushed from 17.31\%~\cite{wah2011caltech} to 85.5\%~\cite{liu2016localizing} within five years.

\begin{figure}
\begin{center}
\includegraphics[width=0.9\linewidth]{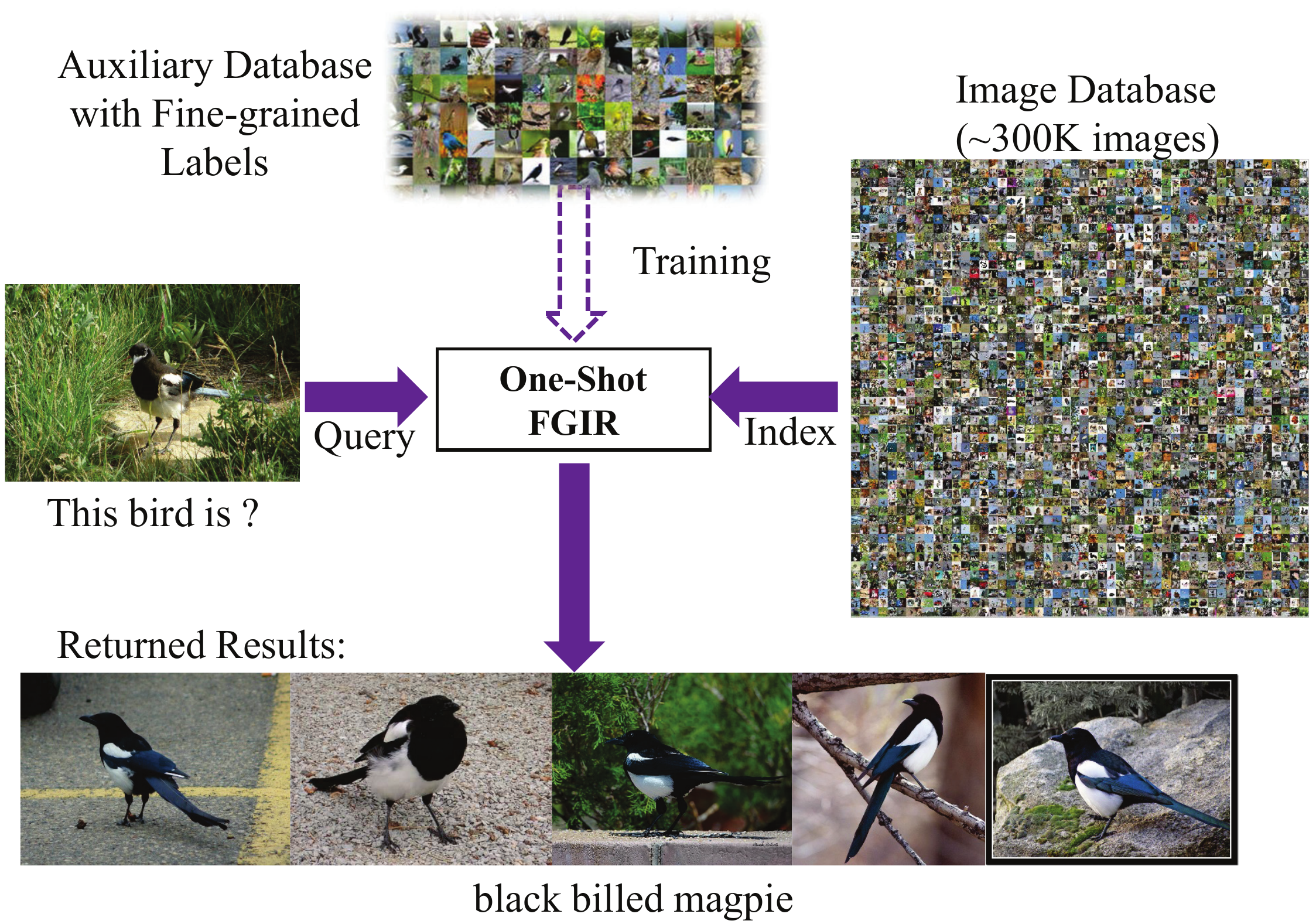}
\end{center}
\caption{Illustration of One-Shot FGIR, which uses prior knowledge inferred from a small independent auxiliary dataset (dashed arrow) to perform fine-grained query of an unseen instance from a large-scale database (solid arrow).}
\vspace{-1.0em}
\label{Fig:FGIR_framework}
\end{figure}

As a special case of visual categorization, FGVC is designed to identify the species existing in the training set. However, the number of fine-grained species in real world could be huge and varying, \emph{e.g.}, new shoes are being designed and produced every week, making it difficult to get a complete training set to recognize unseen objects under the FGVC framework. In other words, FGVC is powerful for scenarios like passenger plane classification, where the number of species is small and the complete training set is easy to acquire. Therefore, it is still an open issue how to perform fine-grained identification without a complete training set.

Motivated to conquer this issue, we present a novel problem named One-Shot Fine-Grained Instance Retrieval (OSFGIR), where ``One-Shot'' emphasizes the ability of identifying the unseen fine-grained species through a retrieval task. As illustrated in Fig.~\ref{Fig:FGIR_framework}, OSFGIR takes an image as input, then performs fine-grained instance retrieval within a large-scale dataset without category labels, and finally returns images containing the identical object. OSFGIR can be tackled by learning powerful features and retrieval models. It is not restricted by the number of learned classifiers, thus has potential to show better generalization ability to unseen species than the FGVC framework. To facilitate model and feature learning, we introduce a small independent auxiliary training set. This incomplete auxiliary set is labeled with fine-grained species and is easy to collect. We thus call this task as One-Shot FGIR, because the training set is small and independent with the testing set.

In this paper, we firstly give the detailed description of OSFGIR, and introduce the OSFGIR-378K dataset. To extract a powerful image feature, we then propose a deep model called Convolutional and Normalization Networks (CN-Nets), which learns and combines two complementary features to generate the object description. Finally, we present a coarse-to-fine OSFGIR framework that consists of coarse retrieval, fine-grained retrieval, and query expansion, respectively. Given a query image, the coarse retrieval firstly returns the Top-K similar images based on a compact descriptor. A more powerful descriptor is hence extracted to rank the Top-K images. Finally, query expansion is used to further improve the retrieval performance. Experimental results show that our feature and retrieval approach significantly outperform existing deep features and retrieval methods in the aspects of both accuracy and efficiency.

OSFGIR is different from and substantially more challenging than traditional instance retrieval task. Most of the instance retrieval datasets are designed for partial-duplicate or semantic-similar search tasks, \emph{e.g.,} Oxford5K~\cite{Philbin07} and Holidays~\cite{jegou2008hamming}. Such problems could be effectively solved by extracting and matching robust local features, \emph{i.e.,} Scale-invariant feature transform (SIFT)~\cite{lowe2004distinctive}, or extracting semantic features by off-the-shelf deep leaning models~\cite{sharif2014cnn}. OSFGIR aims to return images containing the identical fine-grained specie in the query, \emph{e.g.}, images of ``black billed magpie'' with different poses, sizes, backgrounds, \emph{etc}. For OSFGIR, more powerful features are required to identify the fine-grained details among species, because different species may exhibit similar appearances and semantics.

OSFGIR is also a novel retrieval task different from most of existing FGVC works. The most related FGVC work is~\cite{xie2015fine}, where Xie~\emph{et al.} present a fine-grained image search algorithm. Similar to FGVC pipeline, Xie~\emph{et al.}\cite{xie2015fine} learns a series of classifiers based on the training set, and then identifies the fine-grained species with the learned classifiers. Note that, the training and testing datasets share the same species in~\cite{xie2015fine}. OSFGIR differs from~\cite{xie2015fine} in that, it uses independent training and testing datasets, \emph{i.e.}, uses a small incomplete auxiliary dataset for training, but a large-scale dataset for retrieval, which corresponds to more realistic settings.

For the past several years, lots of FGVC works~\cite{zhang2014part,simon2015neural,lin2015deep,liu2016localizing,huang2015part,zhangspda} have been proposed, and they focus on generating image representations from object parts. However, these representations are either complex or require expensive part annotations. CN-Nets is proposed with the motivation of designing a concise representation easy to implement and repeat. It reveals the shortcomings of CNN in feature learning and significantly outperforms the latest deep models in the aspects of efficiency, training complexity, and classification accuracy. FGVC framework is difficult to recognize unseen fine-grained species. This paper defines the OSFGIR problem and presents the OSFGIR-378K dataset. Compared with FGVC, OSFGIR is shown as a better solution for large-scale fine-grained object identification, \emph{e.g.}, our method significantly outperforms recent FGVC works on OSFGIR-378K by more than 11\% on Mean Average Precision. We will release the OSFGIR-378K, and continually enlarge this dataset by adding more species to benefit OSFGIR and large-scale fine-grained object recognition research.


\section{Related Work}
OSFGIR is related to works on fine-grained visual categorization~\cite{berg2013poof,branson2014bird,chai2013symbiotic, gavves2014local, liu2013bird, liu2014part, xiao2014application, zhang2014part,simon2015neural,lin2015deep,ge2015subset,krause2015fine,gavves2013fine,wei2016mask,liu2016localizing,huang2015part,zhangspda} and deep learning-based visual retrieval~\cite{salvador2016faster, wang2014deep,yue2015exploiting,gong2014multi,gordo2016deep,kalantidis2015cross,arandjelovic2015netvlad}. In the following, we summarize these two categories of works respectively.

\emph{Fine-Grained Visual Categorization:} In the past five years, researchers have significantly boosted the classification accuracy of FGVC. Existing methods could be summarized into four categories according to the type of image representation they use, \emph{i.e.,} 1) part-based methods, 2) attribute-based methods, 3) object-based methods, and 4) global-description based methods. 1) As the CUB-200-2011 dataset provides 15 part annotations, the authors of~\cite{berg2013poof} employ the labeled part annotations for training and testing to generate the part description. Based on the labeled part annotations for training images, the other works~\cite{branson2014bird,liu2013bird,liu2014part,zhang2014part,lin2015deep,huang2015part,zhangspda} firstly infer the part annotations for testing images, then generate the part descriptions. As most fine-grained datasets lack manually labeled part annotations, some works infer part labels with unsupervised methods~\cite{gavves2014local,simon2015neural,krause2015fine}. 2) Recently, Liu~\emph{et al.}~\cite{liu2016localizing} employ the given attributes of each part to infer the part annotations, which are then used to generate the object description. 3) Besides the descriptions from local parts, the description from the object bounding box is also commonly used to identify the fine-grained species. \cite{zhang2014part} and~\cite{krause2015fine} infer the bounding boxes for testing images based on those of training images.~\cite{xiao2014application} and~\cite{simon2015neural} generate bounding boxes for training and testing images only with image-level labels. 4) Different from the methods mentioned above, the Bilinear CNN~\cite{lin2015bilinear} and Spatial Transformer Networks (STN)~\cite{jaderberg2015spatial} generate a robust global description for FGVC with a forward pass of the CNN.

\emph{CNN for Visual Retrieval:} CNN has exhibited promising performance for various vision tasks. Several works have attempted to apply CNN in image and instance retrieval~\cite{salvador2016faster,babenko2014neural,yue2015exploiting,gong2014multi,gordo2016deep,kalantidis2015cross,tolias2015particular}. NeuralCode~\cite{babenko2014neural} is an early work that applies CNN for image retrieval, \emph{e.g.,} Babenko~\emph{et al.} employ the output of fully-connected layer as image feature for retrieval. Since Vector Locally Aggregated Descriptors (VLAD)~\cite{jegou2012aggregating} shows good retrieval performance by encoding SIFT descriptors. Ng~\emph{et al.}~\cite{yue2015exploiting} replace the SIFT with CNN feature and encode the convolutional feature maps into a global feature with VLAD. In~\cite{tolias2015particular}, Tolias~\emph{et al.} demonstrate that simply applying a spatial max-pooling over all locations on convolutional feature maps produces an effective visual descriptor. Instance retrieval differs slightly from image retrieval, because it focuses on image regions containing the target object, rather than the entire image. Given the object bounding boxes of query images,  Tolias~\emph{et al.} \cite{tolias2015particular} propose approximate integral max-pooling to select the best matching bounding box from hundreds of candidates. Different from~\cite{tolias2015particular}, Salvador~\emph{et al.}~\cite{salvador2016faster} and Gordo~\emph{et al.}~\cite{gordo2016deep} apply Faster R-CNN~\cite{ren2015faster} to reduce the number of candidate proposals.

OSFGIR differs from FGVC because it is a retrieval task, thus is able to query and identify the unseen query object. OSFGIR is also different from most of the visual retrieval tasks because it needs to further identify and capture the subtle differences among visually and semantically similar objects. Among recent visual retrieval methods, Gordo~\emph{et al.}~\cite{gordo2016deep} have achieve promising performance. However, the method in \cite{gordo2016deep} is not suitable for OSFGIR because: 1) it works on partial-duplicate image retrieval and is evaluated on the widely-used Oxford5K~\cite{Philbin07} and Holidays~\cite{jegou2008hamming}. OSFGIR aims to return images containing the identical fine-grained specie in the query. Those two problems are quite different. 2) The deep regional feature training in~\cite{gordo2016deep} involves keypoint matching to generate the bounding boxes for each candidate object, thus is more suited to partial-duplicate image search.

In the next section, we proceed to give the formulation of OSFGIR, then introduce the OSFGIR-378K dataset.

\section{Problem Formulation}

\subsection{One-Shot Fine-Grained Instance Retrieval}

As illustrated in Fig.~\ref{Fig:FGIR_framework}, OSFGIR defines a fine-grained instance retrieval task assisted with a small independent training set. We denote the set of query images as $\mathcal{Q}=\{{q}_{1},......,{q}_{n}\}$, where $n$ is the number of query images. Each query image has a ground truth label $p_{c} (0\le p_c\le \mathcal{P}_c, 0\le c\le \mathcal{C})$, which denotes the $p$-th fine-grained specie in the $c$-th object category. Note that, we use ``object'' to denote the coarse category and ``specie'' to denote the fine-grained specie within a coarse object category. $\mathcal{C}$ thus is the total number of objects in the query set, $\mathcal{P}_c$ is the number of species in the $c$-th object category. We denote the image database as $\mathcal{D}=\{d_{1}, ,......,d_{m}\}$, where each image either contains a specie in one of the $\mathcal{C}$ categories, or could be a distracter for the retrieval.

Given a query $q$, OSFGIR retrieves the specie in $q$ from $\mathcal{D}$, and returns a ranked list of images. If the query specie exists in the database, OSFGIR aims to return images containing the identical specie. For query species do not exist in the database, OSFGIR returns other species with similar appearances and semantics. Because it does not learn a fixed set of classifiers, OSFGIR is potential to show better generalization ability to new species than existing FGVC methods.

OSFGIR is challenging because fine-grained species commonly exhibit subtle inter-class variance and large intra-class variance. To better tackle this task, we introduce an \emph{Auxiliary Database}, which contains a small set of images annotated with fine-grained specie labels. The auxiliary database is defined as $\mathcal{AD}=\{\alpha_{1},......, \alpha_{k}\}$ and each image $\alpha$ is annotated with a specie label. It allows for feature learning and model fine-tuning, which are potential to significantly improve the OSFGIR performance. Referring to~\cite{fei2006one}, OSFGIR defines an one-shot learning problem, \emph{i.e.}, using prior knowledge in a small independent $\mathcal{AD}$ to identify new objects in the large-scale $\mathcal{D}$. It thus corresponds to more realistic settings than fine-grained image retrieval work~\cite{xie2015fine}.

\subsection{OSFGIR-378K dataset}

\begin{table}
\centering
\small
\caption{The summarization of OSFGIR-378K dataset.}
\begin{tabular}{lrrr}
\hline
Sub-sets & \# Species& \# Queries & \# Images \\
\hline
\hline
 $\mathcal{D}^{bird}$ &200& 1,692& 24,119 \\
 $\mathcal{D}^{car}$ &1,715& 13,033 &136,725 \\
 $\mathcal{D}^{food}$ &70& 7,000 &70,000 \\
 $\mathcal{D}^{distr}$& & & 70,194 \\
\hline
Total ($\mathcal{D}$)&\textbf{1,985} & \textbf{21,725}&\textbf{301,038} \\
\hline
\hline
 $\mathcal{AD}^{bird}$&362&  &30,371 \\
 $\mathcal{AD}^{car}$&196& &16,185 \\
 $\mathcal{AD}^{food}$& 31 & &31,000 \\
\hline
Total ($\mathcal{AD}$)&\textbf{589} &&\textbf{77,556} \\
\hline
\end{tabular}
\label{Tab:database}
\end{table}

Among existing image datasets, ImageNet~\cite{deng2009imagenet} contains many coarse categories like fish, dog, bird, \emph{etc.}, and many fine-grained species. However, ImageNet is designed for image classification and contains a complete training set. Moreover, most of existing baseline deep models are trained on ImageNet, making ImageNet not suitable to serve as a fair benchmark for OSFGIR features and models. Therefore, we collect a new OSFGIR-378K dataset.

We aim to build a large-scale OSFGIR dataset labeled with variety types of fine-grained species, \emph{e.g.}, both the man-made and natural objects. To make the dataset collection task feasible, we leverage existing FGVC datasets to construct the new OSFGIR-378K dataset. Specifically, OSFGIR-378K contains three sub-sets of coarse object categories and one sub-set of distractors. We denote the four sub-sets containing birds, food, cars, and distractors as $\mathcal{D}^{bird}$, $\mathcal{D}^{food}$, $\mathcal{D}^{car}$, and $\mathcal{D}^{distr}$, respectively. In the following, we give details about the construction of those sub-sets.

There are two datasets for fine-grained bird categorization, \emph{i.e.}, CUB-200-2011~\cite{wah2011caltech}, and BirdSnap~\cite{berg2014birdsnap}. BirdSnap contains a larger number of images and species, \emph{i.e.,} 500 species, and 49,829 images in BirdSnap vs. 200 species and 11,788 images in CUB-200-2011, respectively. Note that, the BirdSnap and CUB-200-2011 share 138 common species. Because the images of CUB-200-2011 generally have better quality, we firstly include CUB-200-2011 in the bird sub-set. Then, we put the 138 common species in BirdSnap into the bird sub-set. Finally, we manually delete the noisy images and construct the clean bird sub-set $\mathcal{D}^{bird}$. Therefore, the final $\mathcal{D}^{bird}$ contains the CUB-200-2011 and a part of BirdSnap.

For the car sub-set, there also exist two datasets, \emph{i.e.}, Car196~\cite{Krause20133D} consisting of 16,185 images of 196 species, and CompCar~\cite{yang2015large} consisting of 136,725 images of 1,715 species, respectively. As the CompCar contains more images and species than Car196, and it is a clean dataset, we simply treat the CompCar as $\mathcal{D}^{car}$.

There is only one public dataset for fine-grained food categorization, \emph{i.e.,} the food-101~\cite{bossard2014food}. We thus randomly select $70$ species to generate the food sub-set~$\mathcal{D}^{food}$. To test the robustness and efficiency of OSFGIR methods, we further collect 70K images as distractors $\mathcal{D}^{distr}$. The detailed descriptions of OSFGIR-378K are summarized in Table~\ref{Tab:database}.

The Auxiliary Database $\mathcal{AD}$ contains three sub-sets: $\mathcal{AD}^{bird}$, $\mathcal{AD}^{food}$, and $\mathcal{AD}^{car}$, respectively. $\mathcal{AD}^{bird}$ contains the rest species in BirdSnap, thus has no overlap with the species in $\mathcal{D}^{bird}$. For the car dataset, we treat the Car196 as $\mathcal{AD}^{car}$. $70$ species of food-101 are selected as $\mathcal{D}^{food}$. We hence use the rest $31$ species as $\mathcal{AD}^{food}$. The detailed summarization of  auxiliary database is shown in Table~\ref{Tab:database}.

The final OSFGIR-378K dataset contains a dataset $\mathcal{D}$ for retrieval and an Auxiliary Dataset $\mathcal{AD}$ for training. Note that, to test the ability of identifying unseen species and simulate a real experimental setting, $\mathcal{D}$ and $\mathcal{AD}$ do not share common species and $\mathcal{AD}$ is smaller.

\section{Proposed Approach}
\begin{figure}
\begin{center}
\includegraphics[width=0.9\linewidth]{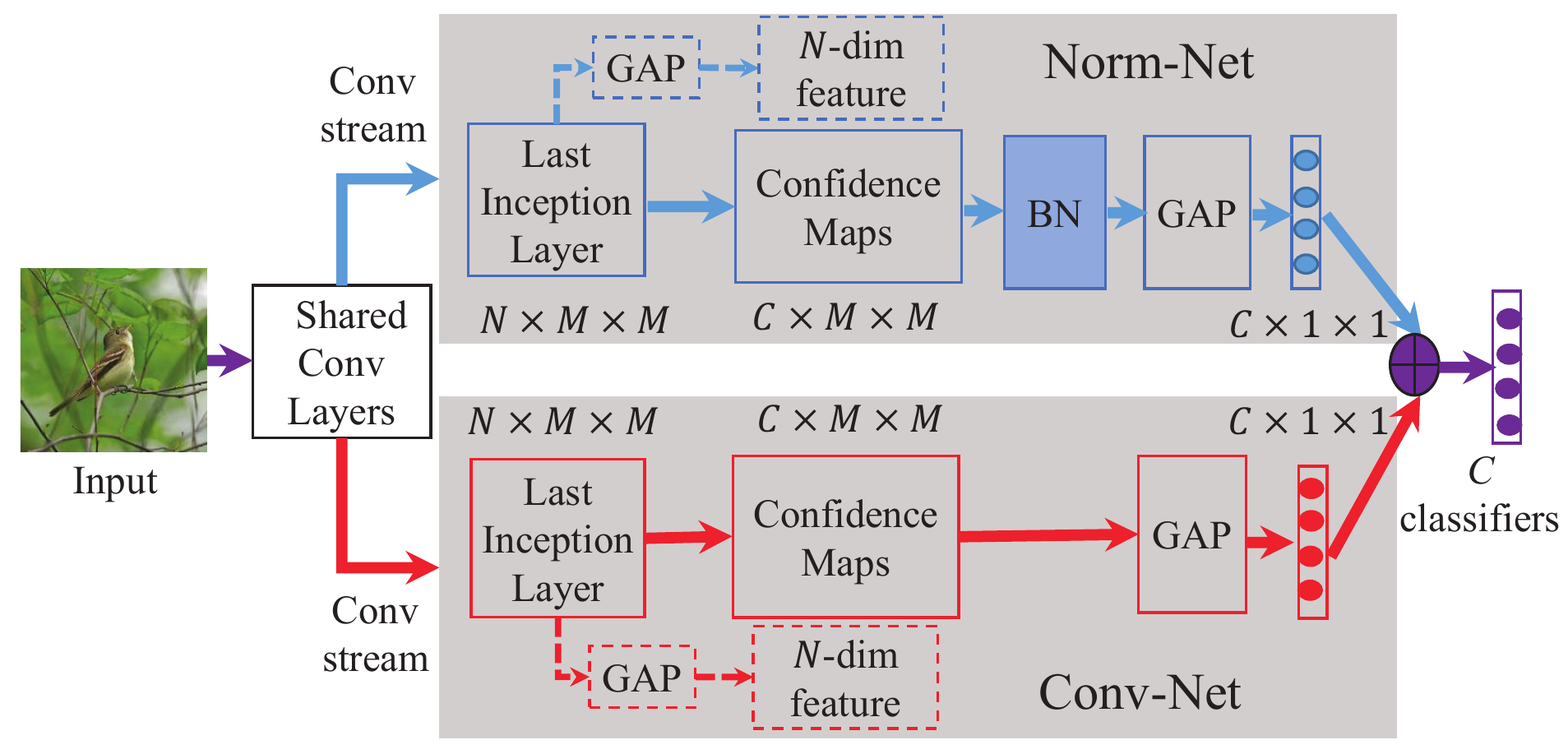}
\end{center}
\caption{Illustration of Convolutional and Normalization Networks (CN-Nets) fine-tuned on the auxiliary dataset. GAP denotes Global Average Pooling and BN denotes the Batch Normalization.}
\label{Fig:network}
\end{figure}

One of the key steps in OSFGIR is to learn a discriminative visual descriptor with auxiliary training set. There exist several CNN-based descriptors that have achieved good classification accuracy on FGVC, \emph{e.g.,} Bilinear CNN~\cite{lin2015bilinear}, Spatial Transformer Networks (STN)~\cite{jaderberg2015spatial}, and CompactBilinear CNN~\cite{gao2015compact}. However, they all have some disadvantages for retrieval task, \emph{e.g.,} time-consuming or hard to extend to unseen data. We propose a novel Convolutional and Normalization Networks (CN-Nets) to generate the image description in Sec.~\ref{sec:cn-nets}. With the CN-Nets, we further propose a coarse-to-fine retrieval framework in Sec.~\ref{sec:retrieval_framework}.

\subsection{Convolutional and Normalization Networks} \label{sec:cn-nets}

CN-Nets is proposed to learn a concise and discriminative representation from image-level labels. It is designed to be more efficient and easy to implement than many FGVC works that generate representations from part labels. As shown in the Fig.~\ref{Fig:network}, CN-Nets takes an image as input and is fine-tuned on the auxiliary dataset in a classification task. It combines outputs from two sub-networks,\emph{ i.e.}, Conv-Net and Norm-Net, as the classification result. Conv-Net and Norm-Net share several convolutional and pooling layers, and are designed with different network structures to learn complementary features. We use outputs of their last inception layers to generate features for OSFGIR. In the following, we introduce these two networks and discuss why their features are complementary to each other.

Most of popular networks, such as Alexnet~\cite{krizhevsky2012imagenet}, VGG~\cite{simonyan2014very}, and GoogLeNet~\cite{szegedy2014going} feed the extracted feature into the fully-connected layer followed by softmax layer for classification. This setting is proven effective in classification tasks but is hard to interpret and is expensive for training due to the huge number of parameters in fully connected layers. Inspired by~\cite{lin2013network}, we propose Conv-Net, which firstly uses convolutional layers to generate feature maps explicitly corresponding to object categories, then uses Global Average Pooling (GAP) layer to predict the classification score for each category. As shown in Fig.~\ref{Fig:network}, Conv-Net firstly generates $C$ feature maps corresponding to $C$ categories, then computes a $C$-dim classification score vector with GAP. Because the average response value on each feature map equals to a classification score, we also call the $C$ feature maps as category confidence maps.

\begin{figure}
\begin{center}
\includegraphics[width=0.9\linewidth]{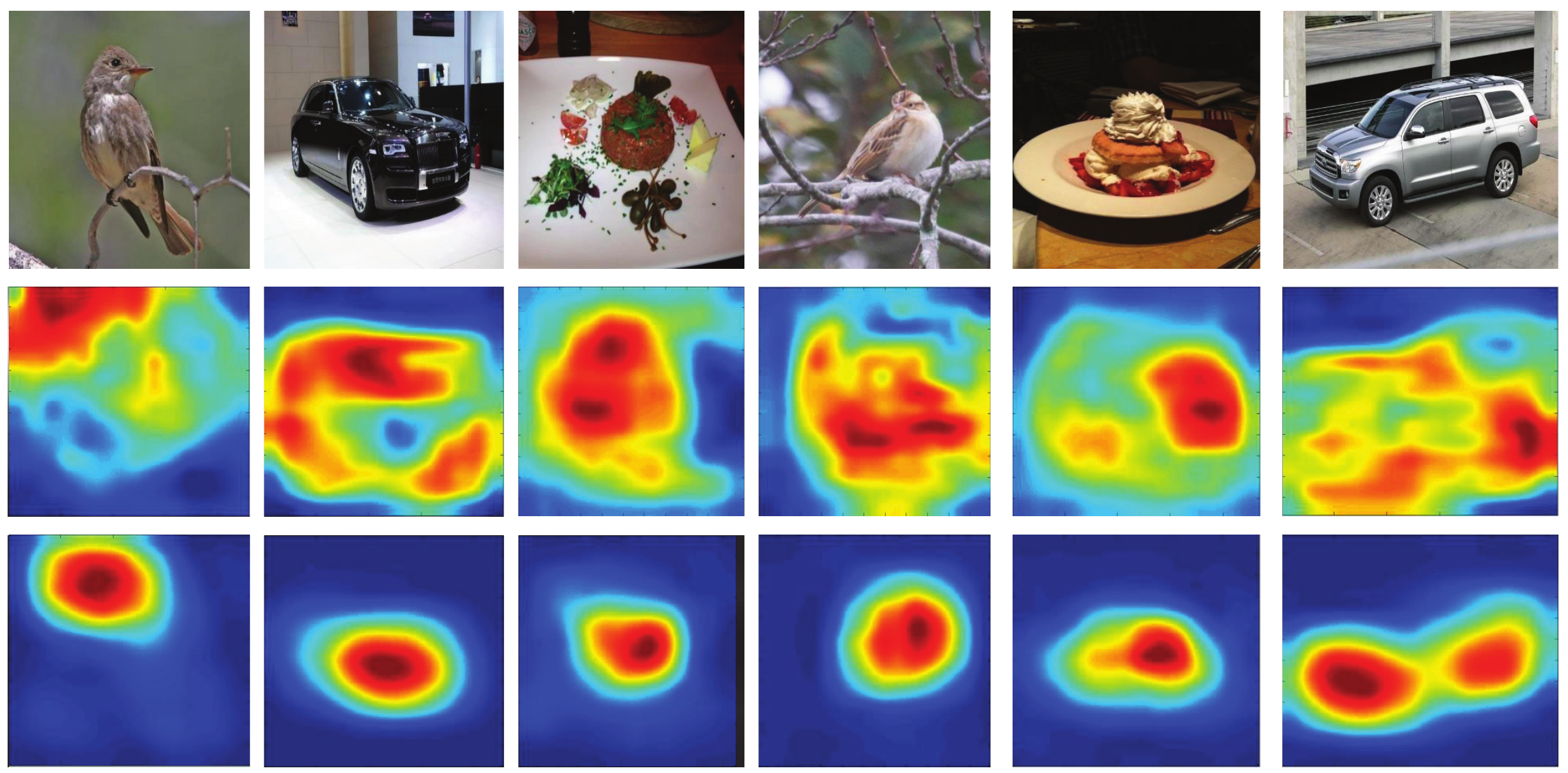}
\end{center}
\caption{Images and their confidence maps generated by Norm-Net (second row) and Conv-Net (third row), respectively. }
\label{Fig:featuremap}
\end{figure}

Compared with the fully-connected layer, GAP layer also generates the classification score and has the following advantages making it more suited for retrieval task: 1) GAP generates explicit object confidence map, \emph{i.e.}, each feature map denotes the spatial activation for an object category. This makes the network feature easier to interpret. 2) GAP has no parameter to tune, thus avoids overfitting and accelerates the network training and testing. 3) The confidence map reveals the discriminative regions in the input image, thus can be useful for object detection and background elimination. As shown in the third row of Fig.~\ref{Fig:featuremap}, the confidence maps of Conv-Net focus on the foreground objects, \emph{i.e.}, the most discriminative regions in the image.

Based on the Conv-Net, we add \emph{Batch Normalization} (BN) layer between the last convolutional layer and GAP layer to construct the Norm-Net. Given $n$ input images in a mini-batch, the BN layer first collects the activations on each location of a $M\times M$ sized feature map as $\mathcal{B}=\{x_{1}, x_{2},......,x_{m}\}$, where $m = n\times M \times M$. BN then employs the mini-batch mean $\mu_{\mathcal{B}}$ and variance $\sigma^{2}_{\mathcal{B}}$ to normalize the samples in $\mathcal{B}$, and finally obtains the normalized values $\widehat{x}$. Aiming to make the output of BN represent the identity transform~\cite{ioffe2015batch} of the input, BN also scales and shifts $\widehat{x}$ by $\gamma$ and $\beta$, respectively. The output of BN $y_{i}$ in Norm-Net is finally passed to the GAP layers to compute the classifier scores. We summarize the BN algorithm in Algorithm~\ref{Alg:bn}. More details of BN can be found in~\cite{ioffe2015batch}.

The second row of Fig.~\ref{Fig:featuremap} shows that the confidence maps of Norm-Net focus on both the foreground object and the spatial contexts, thus are largely different from the confidence maps of Conv-Net. Our experimental results also validate that the features generated by Conv-Net and Norm-Net are complementary to each other. For example, on CUB-200-2011 dataset~\cite{wah2011caltech}, the individual classification accuracies of Conv-Net and Norm-Net features are 83.8\% and 82.3\%, respectively. Combining these two features substantially boosts the accuracy to 85.1\%. More extensive experiments about these two features can be found in Sec.~\ref{sec:exp-cn-nets}.

\begin{algorithm}[t]
\small
\caption{Batch Normalizing Transform, applied to activation $x$ over a mini-batch.}\label{Alg:bn}
\begin{algorithmic}[1]
\STATE{\textbf{Input}: Values of  $x$ over a mini-batch: $\mathcal{B}=\{x_{1}, x_{2},......,x_{m}\}$; Parameters to be learned: $\gamma$, $\beta$}
\STATE{\textbf{Output}: $\{ y_{i}=BN_{\gamma, \beta} (x_{i})\}$}
\STATE{$\mu_{\mathcal{B}}\leftarrow \frac{1}{m}\sum_{i=1}^{m}x_{i}$}
\STATE{$\sigma^{2}_{\mathcal{B}}\leftarrow \frac{1}{m}\sum_{i=1}^{m}(x_{i}-\mu_{\mathcal{B}})^{2}$}
\STATE{$\widehat{x_{i}}\leftarrow \frac{x_{i}-\mu_{\mathcal{B}}}{\sqrt{\sigma^{2}_{\mathcal{B}}+\epsilon}}$}
\STATE{$y_{i}\leftarrow \gamma \widehat{x_{i}}\ + \beta \equiv BN_{\gamma, \beta}(x_{i})$}
\end{algorithmic}
\end{algorithm}

Here, we briefly analyze the reason why the Norm-Net features focus on more spatial contexts. With the BN input $\mathbf{\mathcal{B}}=\{x_{1}, x_{2},......,x_{m}\}$ and output $\{y_{1}, y_{2},......,y_{m}\}$, the backpropagation of the loss $l$, as well as the computation of gradients with respect to the BN parameters can be summarized as, \emph{i.e.},

\begin{equation} \frac{\vartheta{l}}{\vartheta{\widehat{x_{i}}}} = \frac{\vartheta{l}}{\vartheta{y_{i}}} \cdot \gamma ,
\label{eq:bp1}
\end{equation}
\begin{equation}\frac{\vartheta{l}}{\vartheta{\sigma^{2}_{\mathcal{B}}}}=\sum_{i=1}^{m} \frac{\vartheta{l}}{\vartheta{\widehat{x_{i}}}}\cdot (x_{i}-\mu_{\mathcal{B}})\cdot\frac{-1}{2}(\sigma^{2}_{\mathcal{B}}+\epsilon)^{-3/2},
\label{eq:bp2}
\end{equation}
\begin{equation}\frac{\vartheta{l}}{\vartheta{\mu_{\mathcal{B}}}}=\sum_{i=1}^{m} \frac{\vartheta{l}}{\vartheta{\widehat{x_{i}}}}\cdot\frac{-1}{\sqrt{\sigma^{2}_{\mathcal{B}}+\epsilon}}+\frac{\vartheta{l}}{\vartheta{\sigma^{2}_{\mathcal{B}}}}\cdot \frac{\sum_{i=1}^{m}-2(x_{i}-\mu_{\beta})}{m},
\label{eq:bp3}
\end{equation}
\begin{equation}\frac{\vartheta{l}}{\vartheta{x_{i}}} =\frac{\vartheta{l}}{\vartheta{\widehat{x_{i}}}} \cdot \frac{1}{\sqrt{\sigma^{2}_{\mathcal{B}}+\epsilon}} + \frac{\vartheta{l}}{\vartheta{\sigma^{2}_{\mathcal{B}}}}\cdot \frac{2(x_{i}-\mu_{\mathcal{B}})}{m}+\frac{\vartheta{l}}{\vartheta{\mu_{\mathcal{B}}}}\cdot\frac{1}{m},
\label{eq:bp4}
\end{equation}
where Eq.~\eqref{eq:bp4} affects the original activation $x_{i}$. The first and third terms in Eq.~\eqref{eq:bp4} do not affect $x_{i}$ individually. For the first term of Eq.~\eqref{eq:bp4}, because of GAP, $\frac{\vartheta{l}}{\vartheta{y_{i}}}$ is identical for each $x_{i}$ in the same confidence map, so the first term of Eq.~\eqref{eq:bp4} has no effect on $x_{i}$. Among the second term in Eq.~\eqref{eq:bp4}, although the sign for $\frac{\vartheta{l}}{\vartheta{\sigma^{2}_{\mathcal{B}}}}$ is unknown, it has a same effect on all $x_{i}$. Therefore, the second term outputs large values when the gap between $x_{i}$ and $\mu_{\mathcal{B}}$ is large, vice versa. It is easy to infer that, the back propagation would make larger changes to $x_{i}$ in this case and would finally make every $x_{i}$ show similar values close to $\mu_{\mathcal{B}}$. In other words, BN suppresses the highly activated locations on a feature map and encourages the rests.

As a result, Norm-Net tends to activate larger image regions compared with Conv-Net, which only focuses on the discriminative regions on the object. Because the Norm-Net is trained to minimize the classification error, it has potential to discover more helpful contextual cues in the image. Consequently, Conv-Net focuses on the discriminative region, and Norm-Net ``see" more contexts.

By combining features from Conv-Net and Norm-Net, we obtain a more powerful CN-Nets feature. As shown in Fig.~\ref{Fig:network}, given an input image region, Norm-Net and Conv-Net generate the feature maps of the last inception layer, denoted as $\mathcal{X}=\{{\mathcal{X}}_{i}\},i=1,......,\mathcal{N}$, where $\mathcal{N}$=1024 in this paper. The $\mathcal{X}_{i}$ is a 2D tensor denoting the responses of the $i$-th channel. Conv-Net and Norm-Net then generate two $\mathcal{N}$-dim feature vectors with GAP operation. We denote the features of Conv-Net and Norm-Net as $\mathbf{f}_{Conv}$ and $\mathbf{f}_{Norm}$, respectively. The final CN-Nets feature $\mathbf{f}_{CN}$ is generated by concatenating the Conv-Net feature and Norm-Net feature, \emph{i.e.},
\begin{equation}
\mathbf{f}_{CN} = [\mathbf{f}_{Conv}, \mathbf{f}_{Norm}], \mathbf{f}_{CN} \in \mathbf{R}^{2048}.
\end{equation}

\subsection{Coarse-to-fine retrieval framework} \label{sec:retrieval_framework}

\begin{figure}[t]
\begin{center}
\includegraphics[width=0.95\linewidth]{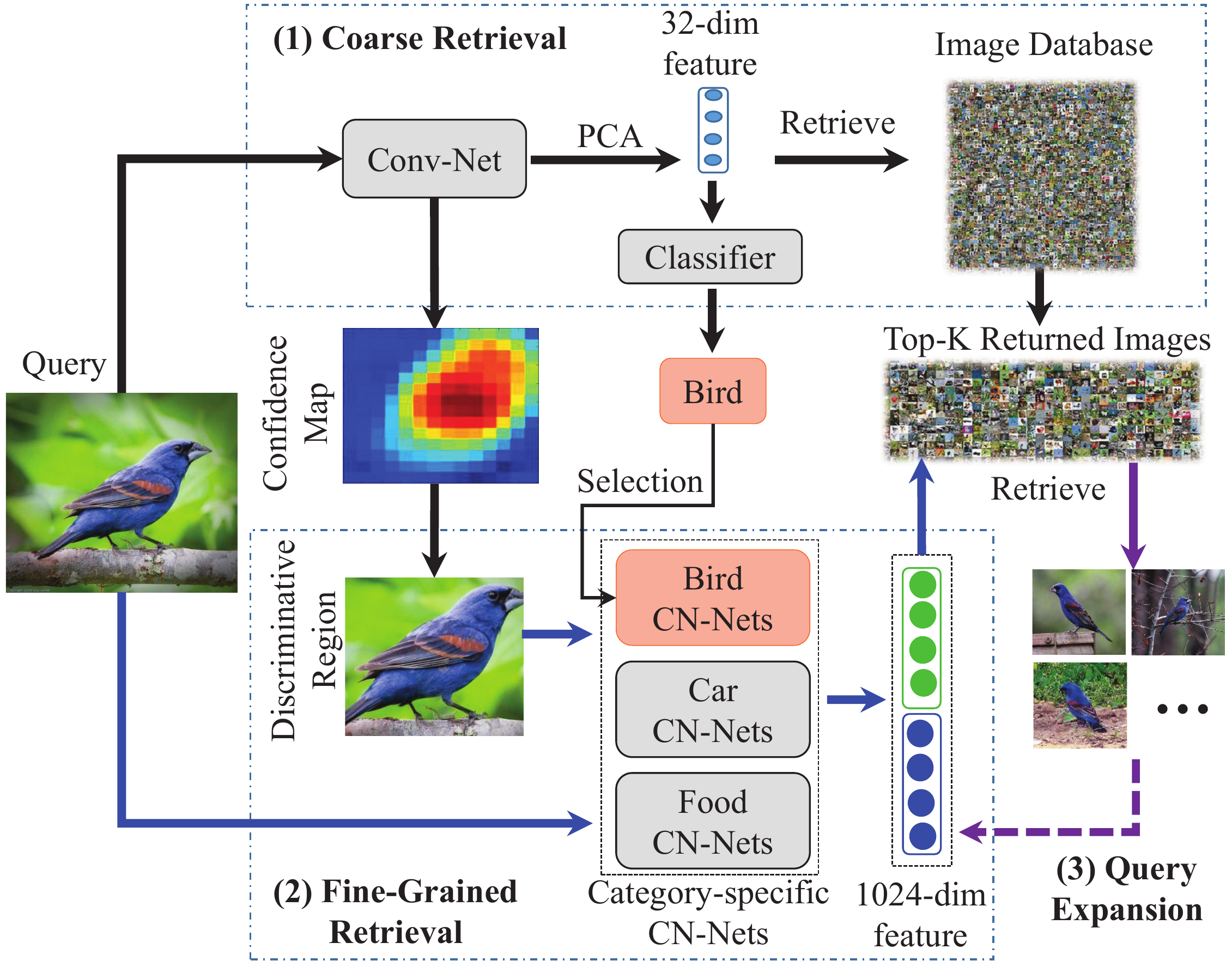}
\end{center}
\caption{Our coarse-to-fine framework for online OSFGIR.}
\vspace{-1.0em}
\label{Fig:framework}
\end{figure}

The fine-grained species in image database belong to different coarse object categories. This naturally leads to a coarse-to-fine retrieval framework, which first quickly retrieves images with the same coarse object to narrow-down the search space, then finds the fine-grained species. As shown in Fig.~\ref{Fig:framework}, the coarse-to-fine framework consists of three stages, \emph{i.e.}, coarse retrieval, fine-grained retrieval, and query expansion. Both the coarse and fine stages are targeted to build an accurate and efficient retrieval system.

\textbf{Coarse Retrieval} retrieves images containing the same coarse object with the query. Because this is an easier task, we extract a compact efficient feature. Specifically, during off-line indexing, we use Conv-Net with the input size of $224\times 224$ to generate the image feature $\mathbf{f}_{Conv}^d\in \mathbf{R}^{1024}$ for the database image $d$. To accelerate the similarity comparison, we reduce the dimensionality of $\mathbf{f}_{Conv}^d$ to $32$-dim with PCA~\cite{dunteman1989principal}, and apply ${L_2}$-normalization to each feature. Moreover, we generate the confidence map for $d$, which is then used to locate the foreground region. As discussed in Sec.~\ref{sec:cn-nets}, Conv-Net outputs $C \times M \times M$ confidence maps for each input image, where $C$ is the number of species in auxiliary dataset, and $M\times M$ is the size of confidence map. Among the $C$ confidence maps, we select the one with the maximum average activation. The selected confidence map is resized to the same size of the input image, and is normalized by dividing the maximum response value on it. We denote the selected confidence map for database image $d$ as $\mathcal{I}^d$.

During online retrieval, we process the query $q$ in the same way to obtain its feature $\mathbf{f}_{Conv}^{q}$ and confidence map $\mathcal{I}^{q}$. Euclidean distance is then computed between the query feature $\mathbf{f}_{Conv}^{q}$ and all database features to obtain the Top-K similar images, \emph{e.g.}, K=10,000. Besides that, we estimate the object category of the query specie. For OSFGIR-378K dataset, a three-way SVM classifier is trained with the $32$-dim feature based on the auxiliary database. We assume that the query images do not contain distractors, thus use a three-way classifier and ignore the distractors. As shown in Fig.~\ref{Fig:framework}, the coarse retrieval outputs 1) the Top-K similar images, 2) the confidence map, and 3) the coarse category label of the query image.

\textbf{Fine-Grained Retrieval} performs fine-grained identification on the Top-K images. As illustrated in Fig.~\ref{Fig:framework}, we employ the complete CN-Nets to extract features $\mathbf{f}_{CN}$ and additionally fuse features from both the discriminative region and the entire image to acquire more discriminative power. The confidence map $\mathcal{I}$ guides the discriminative region selection. $\mathcal{I}$ is first converted to a binary image with threshold $t$,\emph{ e.g.}, $t=0.5$. Then, we employ the connected region analysis~\cite{di1999simple} to remove the small regions and select one dominant region. The discriminative region is generated by cropping the dominant region with the minimum enclosing rectangle. As illustrated in Fig.~\ref{Fig:framework}, the discriminative region covers most of the foreground object.

During online retrieval, we first extract the discriminative regions of query and database images, then extract their CN-Nets features. To ensure the discriminative power of CN-Nets feature, we fine-tune a category-specific CN-Nets for each coarse category on the auxiliary dataset. For instance, if an image is identified to bird category, we employ the CN-Nets fine-tuned on $\mathcal{AD}^{bird}$ for feature extraction. The predicted category label in coarse retrieval hence selects the proper category-specific CN-Nets for feature extraction. This strategy leads to two feature vectors, \emph{i.e.}, $\mathbf{f}_{CN}^{img}\in \mathbf{R}^{2048}$ on the entire image and $\mathbf{f}_{CN}^{reg}\in \mathbf{R}^{2048}$ on the discriminative region, respectively.

We reduce dimensionality of $\mathbf{f}_{CN}^{img}$ and $\mathbf{f}_{CN}^{reg}$ to $512$-dim with PCA, respectively. The final feature $\mathbf{f}$ for fine-grained retrieval is generated by concatenating these two $512$-dim features, \emph{i.e.},
\begin{equation}
\mathbf{f} = [\mathbf{f}_{CN}^{img}, \mathbf{f}_{CN}^{reg}], \mathbf{f} \in \mathbf{R}^{1024}.
\end{equation}
The Top-K images are finally ranked using $\mathbf{f}$ and Euclidean distance. Note that, in our implementation, the discriminative regions and features of database images can be extracted and stored off-line, thus their computations do not degrade the online efficiency.

\textbf{Query Expansion (QE)} strategy is designed to further improve the retrieval accuracy. The fine-grained retrieval effectively ranks some positive images at the top of returned image list. We simply apply average pooling on the features of Top-$\mathcal{K}$ ($\mathcal{K}$=5) returned images to calculate a new descriptor. A new round of retrieval is performed with the new descriptor to update the original ranking list. In the experimental part, we will evaluate the effectiveness and efficiency of this coarse-to-fine retrieval framework.

\section{Experiments}

\subsection{Implementation Details}

We use Caffe~\cite{jia2014caffe} for CNN model training and fine-tuning. The CN-Nets are firstly initialized with the model introduced in~\cite{lim0606}, then are modified based on the Batch-Normalized Convolutional Networks described in ~\cite{ioffe2015batch}. Conv-Net in coarse retrieval is fine-tuned on the complete $\mathcal{AD}$ with $C$=589. The three category-specific CN-Nets in fine-grained retrieval are fine-tuned on $\mathcal{AD}^{bird}$, $\mathcal{AD}^{food}$, and $\mathcal{AD}^{car}$ with $C =$ 362, 196, and 31, respectively. All experiments are conducted on a server equipped with Intel Xeon E5-2650 CPU and  Tesla K40 GPU. Experiments in Sec.~\ref{sec:exp-cn-nets} are conducted on CUB-200-2011~\cite{wah2011caltech}.

We use the widely-used Mean Average Precision (MAP) to evaluate the performance on OSFGIR-378K. Classification accuracy is used when we test on classification tasks.

\subsection{Analysis and Discussions on CN-Nets} \label{sec:exp-cn-nets}

\begin{figure}
\begin{center}
\includegraphics[width=0.8\linewidth,height=3cm]{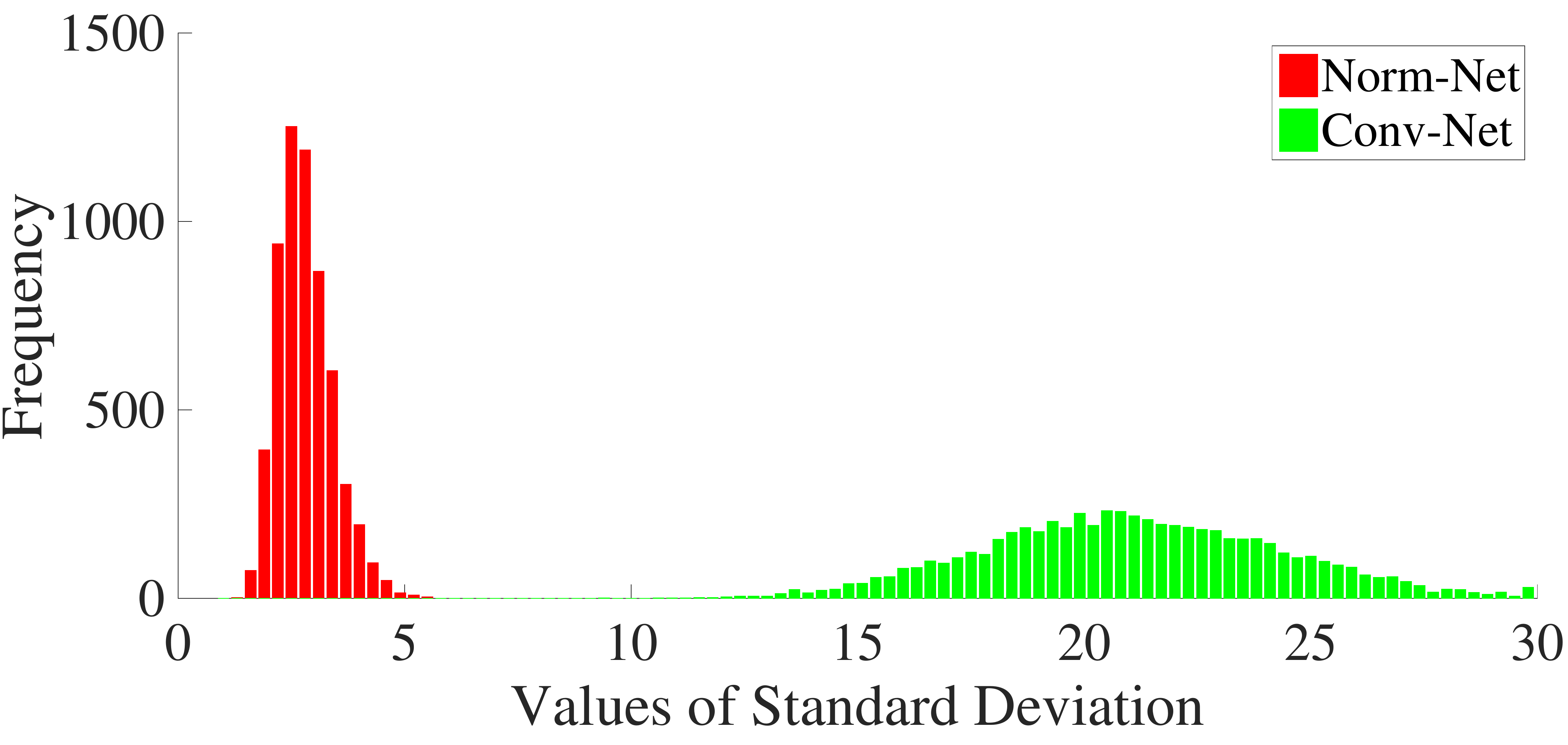}
\end{center}
\caption{Standard deviation histograms of response values in confidence maps produced by Norm-Net and Conv-Net on all training images, respectively.}
\vspace{-0.5em}
\label{Fig:variance}
\end{figure}

Examples in Fig.~\ref{Fig:featuremap} demonstrate that Conv-Net focuses on the discriminative object regions, while Norm-Net covers more spatial contexts. To verify this observation, we calculate the standard deviation (std) histogram of the response values in confidence maps on all training images, and compare the results of Conv-Net and Norm-Net in Fig.~\ref{Fig:variance}. Higher standard deviation (std) denotes more unbalanced activations, \emph{i.e.}, the size of highly activated regions would be smaller, and lower standard deviation (std) denotes that the confidence map contains similar activation values. As shown in the figure, the confidence maps of Conv-Net show substantially larger standard deviations than those of Norm-Net. This means that Conv-Net tends to activate smaller discriminative regions, while Norm-Net covers larger regions.

\begin{figure}
\begin{center}
\includegraphics[width=1\linewidth]{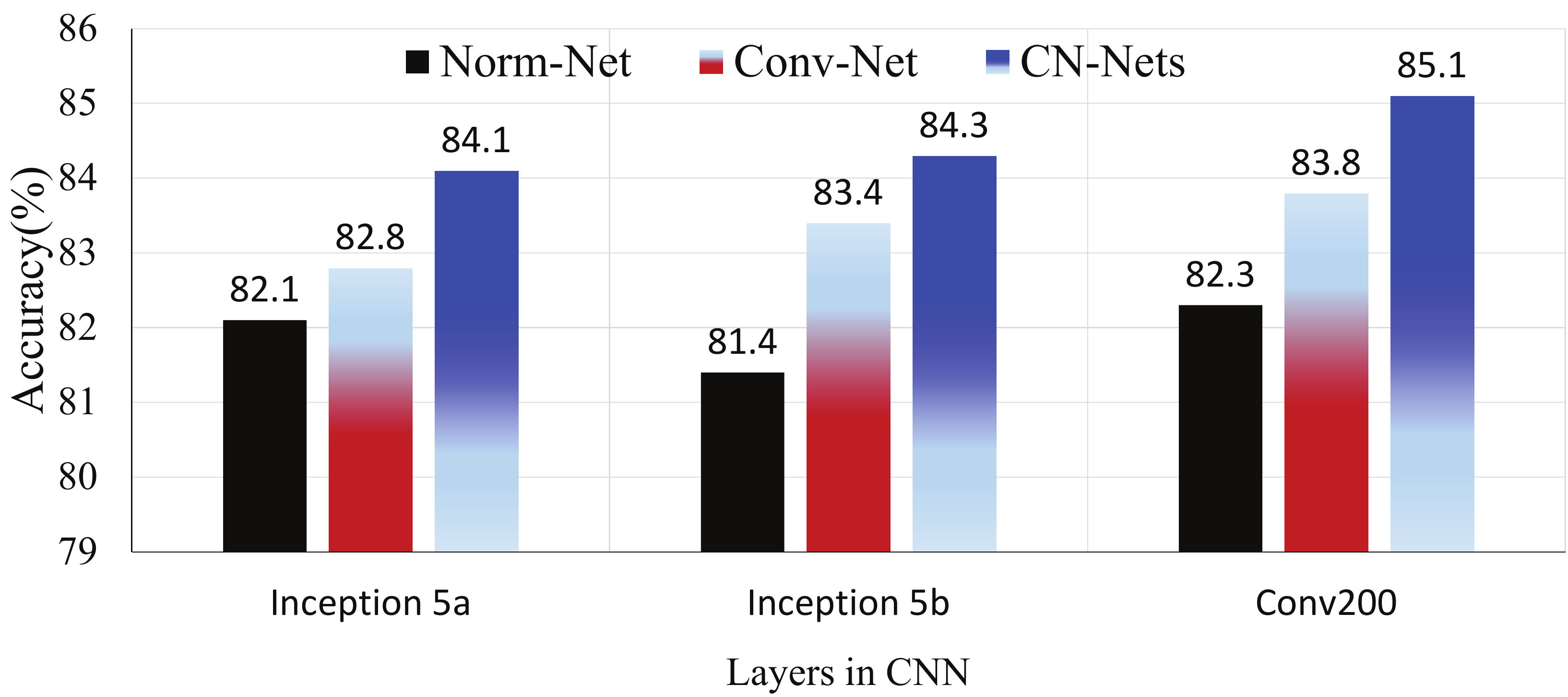}
\end{center}
\caption{The classification accuracy of features extracted by Conv-Net, Norm-Net, and CN-Nets, respectively. Three groups of features are extracted on ``Inception5a", ``Inception5b",``conv200" layers, respectively. Details of these layers can be found in~\cite{szegedy2014going}.}
\vspace{-1.0em}
\label{Fig:compelmentary}
\end{figure}

We conduct another classification experiment to show the complementarity of Conv-Net and Norm-Net features. It is easy to observe from the results in Fig.~\ref{Fig:compelmentary} that, Conv-Net features get higher accuracies than Norm-Net features because they are extracted from more discriminative regions. It is also clear that, the combined Conv-Net and Norm-Net feature, \emph{i.e.}, the CN-Nets feature, gets the best accuracy. The figure also shows that, although BN layer is inserted at the end of Norm-Net, it affects the learned features in preceding layers, \emph{e.g.}, inception5a and 5b.

As shown in Fig.~\ref{Fig:network}, Conv-Net and Norm-Net share several layers in CN-Nets. Table~\ref{Tab:effect_shared_layers} shows the effects of shared layers on the discriminative power and extraction time of CN-Nets feature. It can be observed that, sharing more layers between Conv-Net and Norm-Net actually improves both the efficiency and discriminative power of the CN-Nets feature. For example, by sharing the $C_{1}$-$I_{3b}$ layers, CN-Nets feature obtains the highest classification accuracy and only needs 34\emph{ms} for feature extraction. This validates that the structure of CN-Nets is efficient and reasonable.

\begin{table}
\centering
\small
\caption{The classification accuracy of features extracted by CN-Nets with different number of shared layers. ``Time'' is the feature extraction time. ``$C_{n}$" and ``$I_{m}$" denote the $n$-th convolutional layer and  $m$-th inception layer, respectively. ``$C_{n}$-$C_{m}$" means $C_{n}$ to $C_{m}$ layers are shared. ``-'' denotes no layer is shared. }
\begin{tabular}{cccccc}
\hline
Shared Layers & - & $C_{1}$-$C_{2}$ & $C_{1}$-$I_{3b}$ & $C_{1}$-$I_{4c}$&$C_{1}$-$I_{4e}$  \\
\hline
 \hline
Acc.(\%)&83.9&84.5& 85.1&83.9&84.4\\
Time($ms$) &50&41&34&29&26\\
\hline
\end{tabular}
\vspace{-1.0em}
\label{Tab:effect_shared_layers}
\end{table}

\subsection{Comparison with Other Deep Features} \label{sec:compare_deep_feature}

To test the discriminative power of CN-Nets feature, we compare it with deep features extracted with recent deep learning models, \emph{i.e.,} Res-152~\cite{he2015deep}, BilinearCNN~\cite{lin2015bilinear}, and Spatial Transformer Networks (STN)~\cite{jaderberg2015spatial} in FGVC tasks on CUB-200-2011~\cite{wah2011caltech} and Car196~\cite{Krause20133D}, respectively. We summarize the results in Table~\ref{Tab:comp_feature}.

Among the compared features, CN-Nets feature achieves the best classification accuracy. \emph{E.g.}, single CN-Nets outperforms both the recent BilinearCNN and STN on CUB-200-2011 and Car196. We also compare with an attribute-based method~\cite{liu2016localizing} and a part-based method~\cite{krause2015fine}, which use extra cues for model training and currently report the highest classification accuracies on CUB-200-2011 and Car196, respectively. As shown in Table~\ref{Tab:comp_feature}, by simply concatenating the features from two CN-Nets, fused CN-Nets outperforms both of these works. The above comparisons clearly show CN-Nets is a powerful feature extractor for fine-grained species.

It is also necessary to point out that, CN-Nets is the fastest deep feature extractor in Table~\ref{Tab:comp_feature}. CN-Nets only needs $34ms$ to extract the feature. The Fused CN-Nets needs about $60ms$. They are both faster that STN and BilinearCNN, which cost about $80ms$ and $100ms$, respectively. CN-Nets is also faster than attribute-based~\cite{liu2016localizing} and part-based~\cite{krause2015fine} methods, which need to firstly localize the parts and then generate the description for each part. Therefore, the CN-Nets is also an efficient deep network.

Finally, CN-Nets is also easier to train on new datasets, because it uses simple network structure and involves fewer parameters by removing the fully connected layers. Consequently, CN-Nets features better scalability than more complicated networks like STN and BilinearCNN. As a consequence, the CN-Nets is better suited for the proposed OSFGIR task.

\begin{table}
\centering
\caption{Comparison of classification accuracy (\%) with other deep features. ``Time'' denotes the feature extraction time. ``Complexity" measures the complexity of deep model training on a new dataset. ``Fused CN-Nets'' concatenates the features of two CN-Nets sharing $C_1$-$I_{3b}$ and $C_1$-$I_{4e}$ layers, respectively. ``CN-Nets'' denotes the network sharing $C_1$-$I_{3b}$ layers (refer to Table~\ref{Tab:effect_shared_layers}). }
\begin{tabular}{lcccc}
\hline
Methods & Time($ms$)&  Complexity & CUB200&Car196 \\
\hline
\hline
Res-152~\cite{he2015deep} &120& Easy& 79.8 &90.45\\
STN~\cite{jaderberg2015spatial} &80 & Hard & 84.1 & - \\
Bilinear~\cite{lin2015bilinear} &100 & Medium & 84.1 & 91.3 \\
Recent Report & $\ge$ 100& & 85.5~\cite{liu2016localizing} & 92.6~\cite{krause2015fine} \\
CN-Nets & 34 & Easy & 85.1 & 92.39 \\
Fused CN-Nets & 60 & Easy & \textbf{85.65} & \textbf{93.06} \\
\hline
\end{tabular}
\label{Tab:comp_feature}
\end{table}

\subsection{OSFGIR Performance}

\begin{table}
\caption{The accuracy of the 3-way classifier with different feature dimensionality.}
\centering
\begin{tabular}{cccccc}
\hline
Dim & 1024 & 512 & 128 & 64 & 32 \\
\hline
\hline
Acc.(\%) &97.22&97.12&97.25&97.56& 97.67\\
\hline
\end{tabular}
\label{Tab:category_classifier}
\end{table}

The coarse retrieval extracts a 32-dim feature and trains a three-way category classifier for category-specific CN-Nets selection. We thus first test the validity of this compact feature and the 3-way classifier. Table~\ref{Tab:category_classifier} shows that, the original 1024-dim Conv-Net feature performs well in identifying the coarse object categories. Further applying PCA to reduce the dimensionality does not degrade the accuracy, \emph{e.g.,} 32-dim feature achieves a slightly better accuracy of 97.67\%. Therefore, the 32-dim feature and the classifier are effective for coarse retrieval and CN-Nets selection.

We then discuss the accuracy of the coarse-to-fine retrieval framework. As shown in Table~\ref{Tab:results1}, coarse retrieval achieves the Mean Average Precision (MAP) of 9.18\%. The coarse-to-fine retrieval with the 2048-dim feature $\mathbf{f}_{CN}^{img}$ significantly boosts the performance from 9.18\% to 22.19\%. After reducing the feature dimensionality to 512, the retrieval performance increases to 22.22\%. The large improvement over coarse retrieval demonstrates the importance of fine-gained retrieval stage. The retrieval results with complete feature $\mathbf{f}$ concatenating $\mathbf{f}_{CN}^{img}$ and $\mathbf{f}_{CN}^{reg}$ will be presented in Sec.~\ref{sec:compare}.

During fine-grained retrieval, we use category-specific CN-Nets for feature extraction. To show the validity of this strategy, we compare with the performance of a Unified-CN-Nets fine-tuned on the complete auxiliary dataset. As shown in Table~\ref{Tab:results1}, category-specific CN-Nets plus the accurate 3-way category classifier achieves a significantly better performance, \emph{e.g.,} 22.19\% \emph{vs} 16.90\% of Unified-CN-Nets. The above experiments clearly show the validity of our coarse-to-fine retrieval framework.

\begin{table}
\centering
\caption{The retrieval performance on OSFGIR-378K. ``Dim'' denotes the feature dimensionality. ``Coarse'' and ``C-to-F'' denote the coarse retrieval and coarse-to-fine retrieval, respectively.}
\begin{tabular}{lccc}
\hline
Methods &Net. input size& Dim &MAP(\%)\\
\hline
\hline
Coarse &224$\times$224 & 1024 &11.27\\
Coarse &224$\times$224 & 32 &9.18\\
C-to-F\_$\mathbf{f}_{CN}^{img}$&448$\times$448 & 2048  & 22.19\\
C-to-F\_$\mathbf{f}_{CN}^{img}$&448$\times$448 & 512  & 22.22\\
C-to-F\_{Unified} &448$\times$448& 2048 &16.90\\
\hline
\end{tabular}
\label{Tab:results1}
\end{table}

The online querying time consists of five operations, \emph{i.e.}, two feature extractions, two retrievals, and one classification. We finally analyze the time complexity and summarize the details in Table~\ref{Tab:time}. As shown in the table, our total online query time is about 270\emph{ms}. The coarse retrieval returns results in less than 100$ms$ to reduce the search space. This allows the online retrieval to finish retrieval in less than 170$ms$ using more powerful features.

\begin{table}
\small
\centering
\caption{The time complexity of our OSFGIR system. $*$ means the running-time is evaluated on GPU K40.}
\begin{tabular}{c|c|c}
\hline
 Stages & Operations & Time($ms$)\\
\hline
Coarse	& Feature Extraction &21$^{*}$\\
Retrieval	& Retrieval & 73.1\\
	& Classifier & 8\\
\hline
Fine-Grained	& Feature Extraction& 68$^{*}$\\
Retrieval	& Retrieval &100\\
\hline
Total & & 270.1\\
\hline
\end{tabular}
\label{Tab:time}
\end{table}

\begin{table}
\centering
\small
\caption{Comparison with other methods on OSFGIR-378K.}
\begin{tabular}{lccc}
\hline
Methods &Net. input size& Dim &MAP(\%)\\
\hline
\hline
VGG19+VLAD~\cite{yue2015exploiting} &224$\times$224& 512&4.14 \\
VGG19+MAC~\cite{tolias2015particular}&224$\times$224 &512&9.6 \\
VGG19+NeuralCode~\cite{babenko2014neural} &224$\times$224& 4096  & 10.17\\
VGG19+CROW~\cite{kalantidis2015cross}&224$\times$224 &512  &10.73 \\
Res-152~\cite{he2015deep}&224$\times$224 &1024 &12.13\\
L2\_${\mathbf{f}_{CN}}$  & 224$\times$224 & 2048 & 12.93 \\
\hline
\hline
GoogLeNet&448$\times$448 &1024  &14.9\\
Res-152&448$\times$448 &1024 & 14.7\\
CompactBilinear CNN~\cite{gao2015compact} &448$\times$448&8192& 15.06\\
L2\_${\mathbf{f}_{CN}}$ & 448$\times$448 & 2048 & 16.90 \\
C-to-F\_$\mathbf{f}$&448$\times$448 & 1024  & \textbf{23.68}\\
C-to-F\_$\mathbf{f}$+QE&448$\times$448 & 1024& \textbf{26.31}\\
\hline
\end{tabular}
\label{Tab:results}
\end{table}
\subsection{Comparison with Existing Methods} \label{sec:compare}

To further test the performance of our OSFGIR system, we compare with recent image and instance retrieval methods including NeuralCode~\cite{babenko2014neural}, CROW~\cite{kalantidis2015cross}, CNN+VLAD~\cite{yue2015exploiting}, MAC~\cite{tolias2015particular}, and CompactBilinear CNN~\cite{gao2015compact}. To make the comparison fair, we fine-tune the network of each method on the auxiliary dataset. For NerualCode, we use the output of the $fc6$ layer in VGG19 as the feature. For the method of~\cite{yue2015exploiting,tolias2015particular}, we employ the output of the last convolutional layer, \emph{i.e.},~\emph{conv5\_4} to extract feature. The CROW is implemented based on the \emph{pool5} of VGG19. The comparsions with different network input sizes are summarized in Table~\ref{Tab:results}.

L2\_${\mathbf{f}_{CN}}$ denotes directly using Euclidean distance and the 2048-D original CN-Nets feature extracted from the entire image for retrieval. The comparison shows our work outperforms existing retrieval methods by large margins, \emph{e.g.,} our work achieves the MAP of 23.68\% using the final 1024-dim CN-Nets feature $\mathbf{f}$. This is significantly better than 14.7\% of Res-152, and 15.06\% of CompactBilinear CNN~\cite{gao2015compact}.

\begin{figure}
\begin{center}
\includegraphics[width=0.75\linewidth]{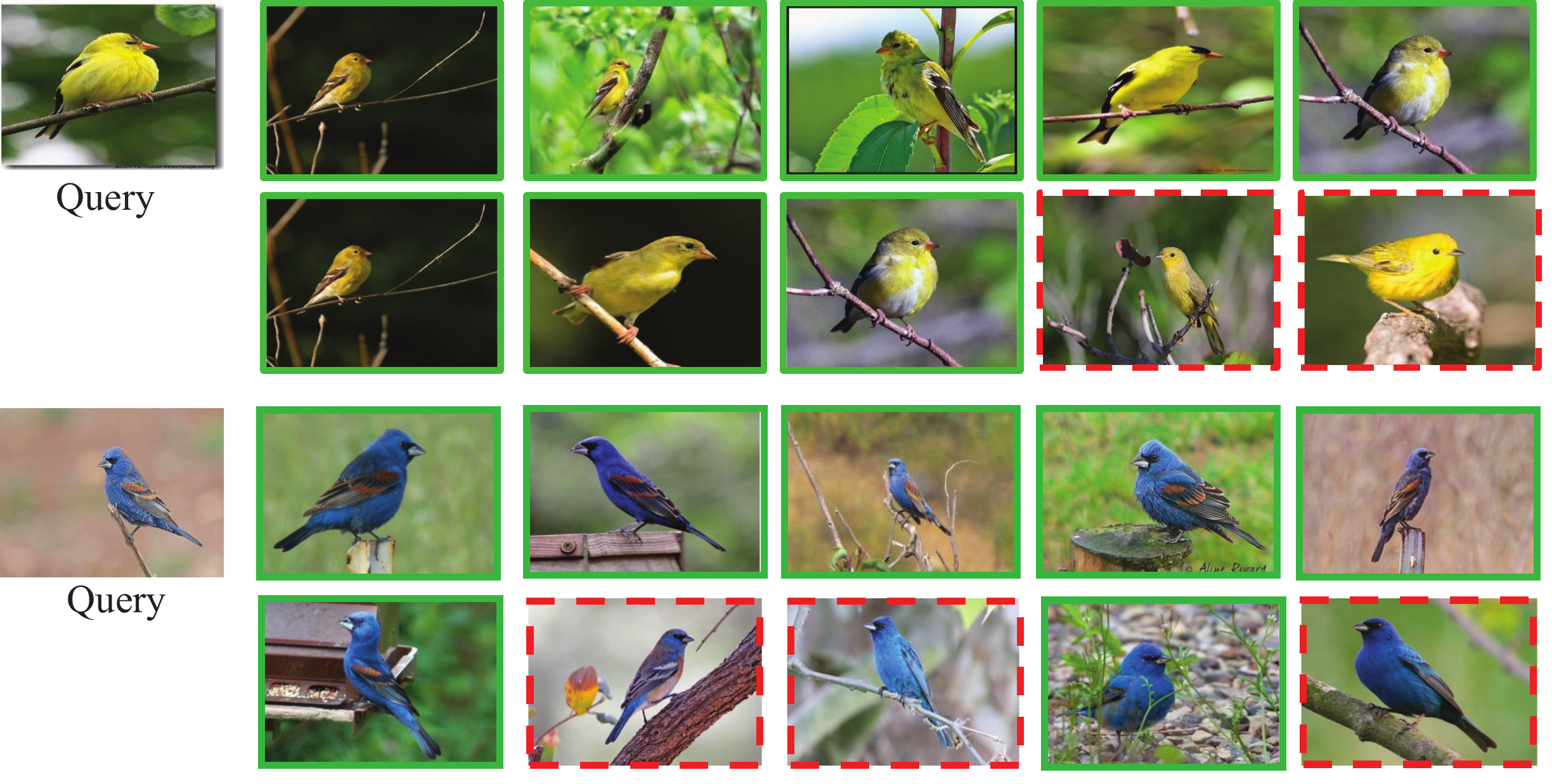}
\includegraphics[width=0.75\linewidth]{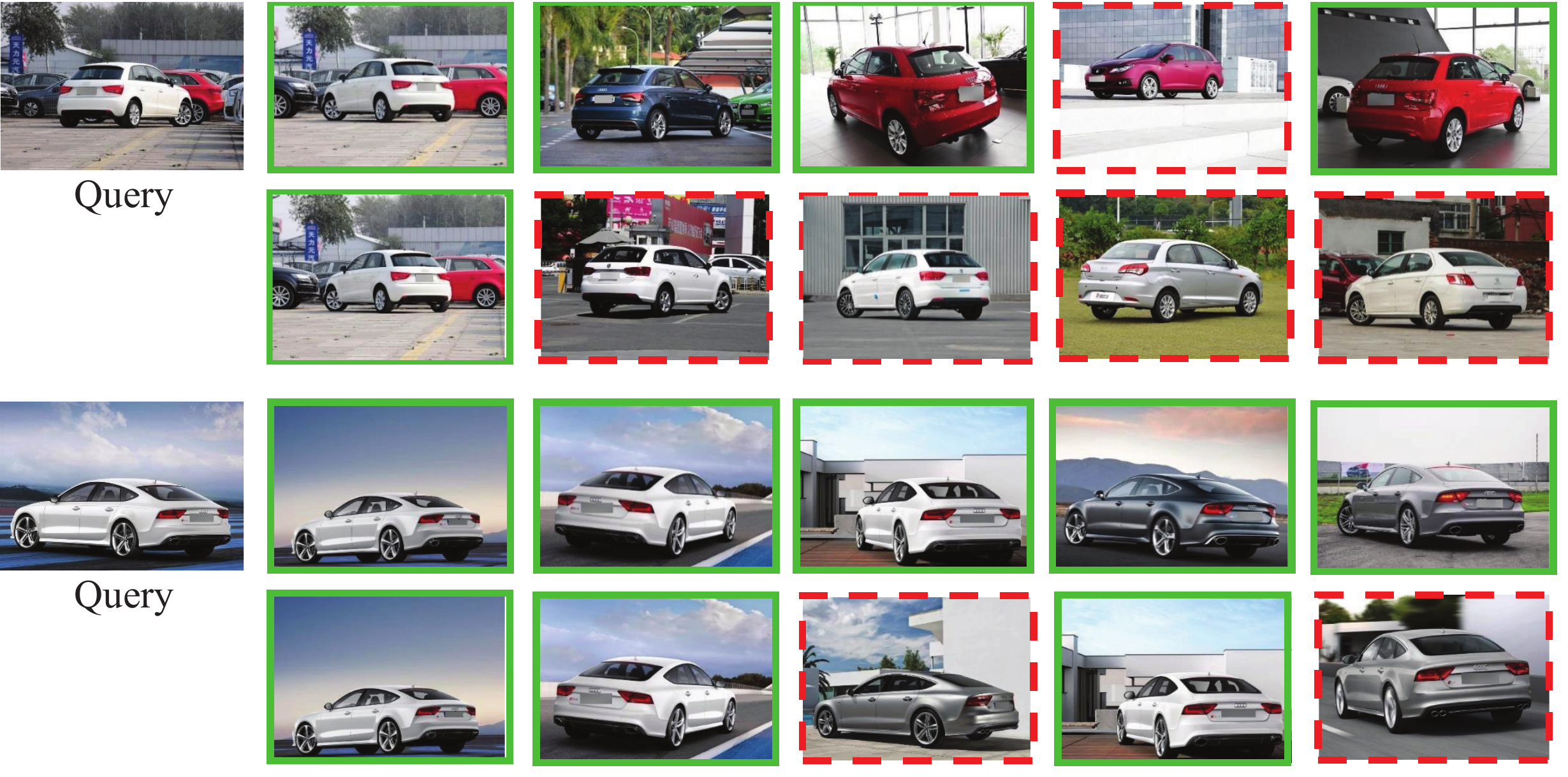}
\includegraphics[width=0.75\linewidth]{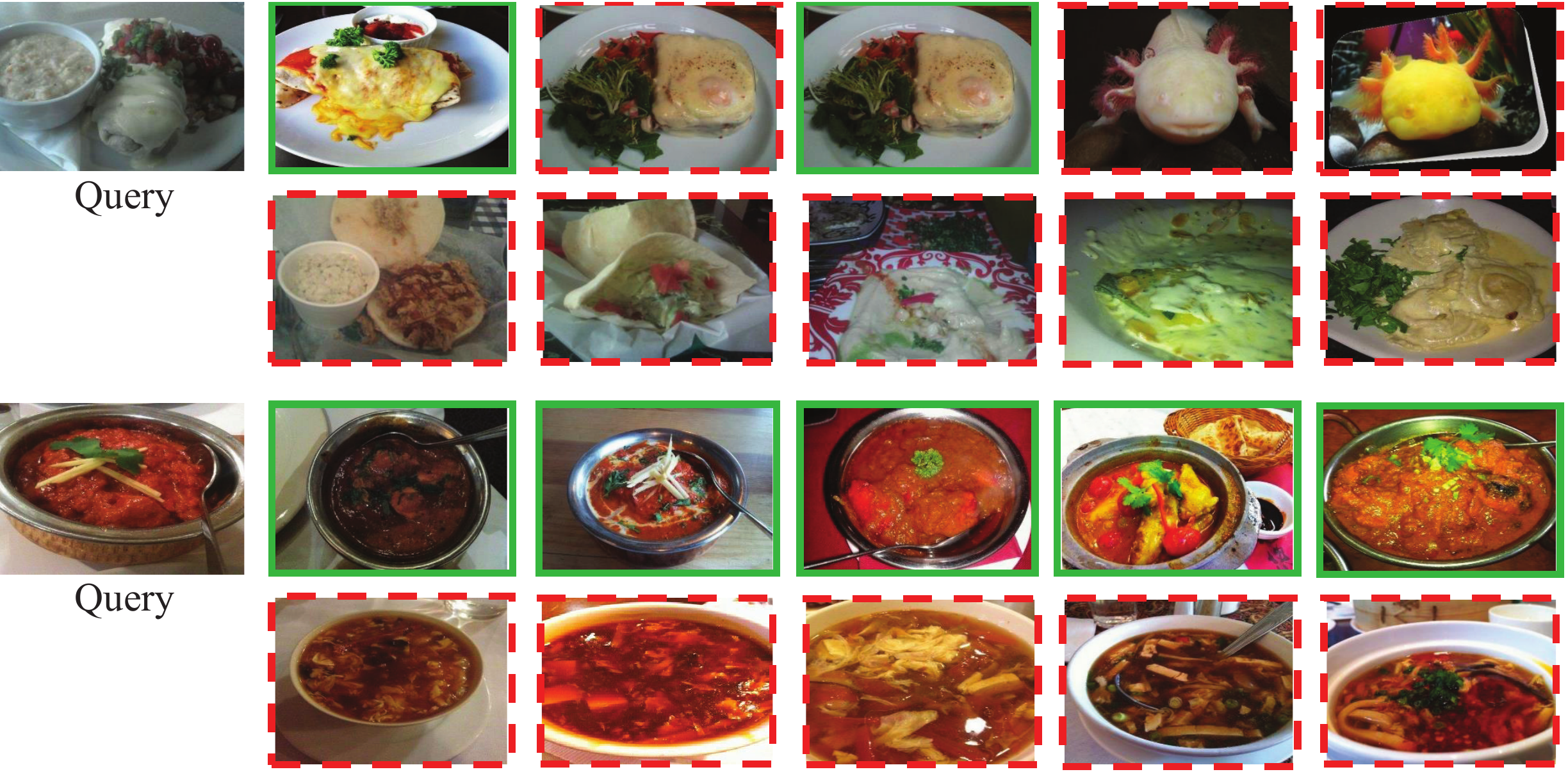}
\end{center}
\caption{Several examples of Top-5 retrieved images of our method (first row) and CompactBilinearCNN~\cite{gao2015compact} (second row). Green solid and red dashed bounding boxes denote true positive and false positive, respectively. }
\label{Fig:ret_result}
\end{figure}

Referring to the results in Table~\ref{Tab:results1}, we can infer that the final 1024-dim CN-Nets feature $\mathbf{f}$ boosts the retrieval performance from 22.22\% to 23.68\%. Additionally, applying QE further improves the retrieval performance to 26.31\%. Therefore, we can conclude that, CN-Nets is a powerful feature extractor, and our coarse-to-fine retrieval work is effective and efficient for OSFGIR. Therefore, this work proposes an effective and efficient solution to the challenge of large-scale fine-grained identification of unseen objects. Retrieval examples of our work can be found in Fig.~\ref{Fig:ret_result}, where our method substantially outperforms the CompactBilinerCNN~\cite{gao2015compact}.

\section{Conclusion}

This paper presents a novel OSFGIR approach to tackle the challenging large-scale fine-grained identification of unseen objects. OSFGIR aims to return a ranked list of images containing the identical fine-grained specie in the query from a large-scale dataset. We first define the OSFGIR problem and construct a OSFGIR-378K dataset containing about 378K images and 1,985 fine-grained species. To extract discriminative descriptors for the fine-grained species, we propose the Convolutional and Normalization Networks (CN-Nets). CN-Nets conducts one-shot learning on the auxiliary dataset to extract two complementary deep features as the representation of target testing data. A coarse-to-fine retrieval framework is hence proposed to chase a reasonable trade-off between efficiency and accuracy. Experiments on OSFGIR-378K show that our descriptor and retrieval framework achieve significantly better performance than existing FGVC and image retrieval methods. Further works will be conducted to explore more efficient one-shot learning algorithms to optimize both the feature extraction and indexing modules.

\newpage
\bibliographystyle{abbrv}
\bibliography{egbib}

\begin{thebibliography}{10}

\bibitem{lim0606}
https://github.com/lim0606/caffe-googlenet-bn.

\bibitem{arandjelovic2015netvlad}
R.~Arandjelovi{\'c}, P.~Gronat, A.~Torii, T.~Pajdla, and J.~Sivic.
\newblock Netvlad: Cnn architecture for weakly supervised place recognition.
\newblock In {\em CVPR}. IEEE, 2016.

\bibitem{babenko2014neural}
A.~Babenko, A.~Slesarev, A.~Chigorin, and V.~Lempitsky.
\newblock Neural codes for image retrieval.
\newblock In {\em ECCV}, pages 584--599. Springer, 2014.

\bibitem{berg2013poof}
T.~Berg and P.~N. Belhumeur.
\newblock Poof: Part-based one-vs.-one features for fine-grained
  categorization, face verification, and attribute estimation.
\newblock In {\em CVPR}, pages 955--962. IEEE, 2013.

\bibitem{berg2014birdsnap}
T.~Berg, J.~Liu, S.~W. Lee, M.~L. Alexander, D.~W. Jacobs, and P.~N. Belhumeur.
\newblock Birdsnap: Large-scale fine-grained visual categorization of birds.
\newblock In {\em CVPR}, pages 2019--2026. IEEE, 2014.

\bibitem{bossard2014food}
L.~Bossard, M.~Guillaumin, and L.~Van~Gool.
\newblock Food-101--mining discriminative components with random forests.
\newblock In {\em ECCV}, pages 446--461. Springer, 2014.

\bibitem{branson2014bird}
S.~Branson, G.~Van~Horn, S.~Belongie, and P.~Perona.
\newblock Bird species categorization using pose normalized deep convolutional
  nets.
\newblock In {\em BMVC}, 2014.

\bibitem{chai2013symbiotic}
Y.~Chai, V.~Lempitsky, and A.~Zisserman.
\newblock Symbiotic segmentation and part localization for fine-grained
  categorization.
\newblock In {\em ICCV}, pages 321--328. IEEE, 2013.

\bibitem{deng2009imagenet}
J.~Deng, W.~Dong, R.~Socher, L.-J. Li, K.~Li, and L.~Fei-Fei.
\newblock Imagenet: A large-scale hierarchical image database.
\newblock In {\em CVPR}, 2009.

\bibitem{di1999simple}
L.~Di~Stefano and A.~Bulgarelli.
\newblock A simple and efficient connected components labeling algorithm.
\newblock In {\em ICIAP}, pages 322--327. IEEE, 1999.

\bibitem{dunteman1989principal}
G.~H. Dunteman.
\newblock {\em Principal components analysis}.
\newblock Number~69. Sage, 1989.

\bibitem{fei2006one}
L.~Fei-Fei, R.~Fergus, and P.~Perona.
\newblock One-shot learning of object categories.
\newblock {\em TPAMI}, 28(4):594--611, 2006.

\bibitem{gao2015compact}
Y.~Gao, O.~Beijbom, N.~Zhang, and T.~Darrell.
\newblock Compact bilinear pooling.
\newblock In {\em CVPR}. IEEE, 2016.

\bibitem{gavves2013fine}
E.~Gavves, B.~Fernando, C.~G. Snoek, A.~W. Smeulders, and T.~Tuytelaars.
\newblock Fine-grained categorization by alignments.
\newblock In {\em ICCV}, pages 1713--1720. IEEE, 2013.

\bibitem{gavves2014local}
E.~Gavves, B.~Fernando, C.~G. Snoek, A.~W. Smeulders, and T.~Tuytelaars.
\newblock Local alignments for fine-grained categorization.
\newblock {\em IJCV}, pages 1--22, 2014.

\bibitem{ge2015subset}
Z.~Ge, C.~McCool, C.~Sanderson, and P.~Corke.
\newblock Subset feature learning for fine-grained category classification.
\newblock In {\em CVPR}, number IEEE, 2015.

\bibitem{gong2014multi}
Y.~Gong, L.~Wang, R.~Guo, and S.~Lazebnik.
\newblock Multi-scale orderless pooling of deep convolutional activation
  features.
\newblock In {\em ECCV}, pages 392--407. Springer, 2014.

\bibitem{gordo2016deep}
A.~Gordo, J.~Almazan, J.~Revaud, and D.~Larlus.
\newblock Deep image retrieval: Learning global representations for image
  search.
\newblock In {\em ECCV}. Springer, 2016.

\bibitem{he2015deep}
K.~He, X.~Zhang, S.~Ren, and J.~Sun.
\newblock Deep residual learning for image recognition.
\newblock In {\em CVPR}. IEEE, 2016.

\bibitem{huang2015part}
S.~Huang, Z.~Xu, D.~Tao, and Y.~Zhang.
\newblock Part-stacked cnn for fine-grained visual categorization.
\newblock In {\em CVPR}. IEEE, 2016.

\bibitem{ioffe2015batch}
S.~Ioffe and C.~Szegedy.
\newblock Batch normalization: Accelerating deep network training by reducing
  internal covariate shift.
\newblock In {\em ICML}, 2015.

\bibitem{jaderberg2015spatial}
M.~Jaderberg, K.~Simonyan, A.~Zisserman, et~al.
\newblock Spatial transformer networks.
\newblock In {\em NIPS}, pages 2008--2016, 2015.

\bibitem{jegou2008hamming}
H.~J{\'e}gou, M.~Douze, and C.~Schmid.
\newblock Hamming embedding and weak geometry consistency for large scale image
  search-extended version.
\newblock 2008.

\bibitem{jegou2012aggregating}
H.~Jegou, F.~Perronnin, M.~Douze, J.~S{\'a}nchez, P.~Perez, and C.~Schmid.
\newblock Aggregating local image descriptors into compact codes.
\newblock {\em TPAMI}, 34(9):1704--1716, 2012.

\bibitem{jia2014caffe}
Y.~Jia, E.~Shelhamer, J.~Donahue, S.~Karayev, J.~Long, R.~Girshick,
  S.~Guadarrama, and T.~Darrell.
\newblock Caffe: Convolutional architecture for fast feature embedding.
\newblock {\em arXiv preprint arXiv:1408.5093}, 2014.

\bibitem{kalantidis2015cross}
Y.~Kalantidis, C.~Mellina, and S.~Osindero.
\newblock Cross-dimensional weighting for aggregated deep convolutional
  features.
\newblock In {\em ECCVW}. Springer, 2016.

\bibitem{krause2015fine}
J.~Krause, H.~Jin, J.~Yang, and L.~Fei-Fei.
\newblock Fine-grained recognition without part annotations.
\newblock In {\em CVPR}, pages 5546--5555. IEEE, 2015.

\bibitem{Krause20133D}
J.~Krause, M.~Stark, J.~Deng, and L.~Fei-Fei.
\newblock 3d object representations for fine-grained categorization.
\newblock In {\em ICCVW}, pages 554--561. IEEE, 2013.

\bibitem{krizhevsky2012imagenet}
A.~Krizhevsky, I.~Sutskever, and G.~E. Hinton.
\newblock Imagenet classification with deep convolutional neural networks.
\newblock In {\em NIPS}, pages 1097--1105, 2012.

\bibitem{lin2015deep}
D.~Lin, X.~Shen, C.~Lu, and J.~Jia.
\newblock Deep lac: Deep localization, alignment and classification for
  fine-grained recognition.
\newblock In {\em CVPR}, pages 1666--1674. IEEE, 2015.

\bibitem{lin2013network}
M.~Lin, Q.~Chen, and S.~Yan.
\newblock Network in network.
\newblock In {\em ICLR}, 2014.

\bibitem{lin2015bilinear}
T.-Y. Lin, A.~RoyChowdhury, and S.~Maji.
\newblock Bilinear cnn models for fine-grained visual recognition.
\newblock In {\em ICCV}. IEEE, 2015.

\bibitem{liu2013bird}
J.~Liu and P.~N. Belhumeur.
\newblock Bird part localization using exemplar-based models with enforced pose
  and subcategory consistency.
\newblock In {\em ICCV}, pages 2520--2527. IEEE, 2013.

\bibitem{liu2014part}
J.~Liu, Y.~Li, and P.~N. Belhumeur.
\newblock Part-pair representation for part localization.
\newblock In {\em ECCV}, pages 456--471. Springer, 2014.

\bibitem{liu2016localizing}
X.~Liu, J.~Wang, S.~Wen, E.~Ding, and Y.~Lin.
\newblock Localizing by describing: Attribute-guided attention localization for
  fine-grained recognition.
\newblock In {\em NIPS}, 2016.

\bibitem{lowe2004distinctive}
D.~G. Lowe.
\newblock Distinctive image features from scale-invariant keypoints.
\newblock {\em IJCV}, 60(2):91--110, 2004.

\bibitem{Philbin07}
J.~Philbin, O.~Chum, M.~Isard, J.~Sivic, and A.~Zisserman.
\newblock Object retrieval with large vocabularies and fast spatial matching.
\newblock In {\em CVPR}. IEEE, 2007.

\bibitem{ren2015faster}
S.~Ren, K.~He, R.~Girshick, and J.~Sun.
\newblock Faster r-cnn: Towards real-time object detection with region proposal
  networks.
\newblock In {\em NIPS}, 2015.

\bibitem{salvador2016faster}
A.~Salvador, X.~Gir{\'o}-i Nieto, F.~Marqu{\'e}s, and S.~Satoh.
\newblock Faster r-cnn features for instance search.
\newblock In {\em CVPRW}. IEEE, 2016.

\bibitem{sharif2014cnn}
A.~Sharif~Razavian, H.~Azizpour, J.~Sullivan, and S.~Carlsson.
\newblock Cnn features off-the-shelf: an astounding baseline for recognition.
\newblock In {\em CVPR}, pages 806--813. IEEE, 2014.

\bibitem{simon2015neural}
M.~Simon and E.~Rodner.
\newblock Neural activation constellations: Unsupervised part model discovery
  with convolutional networks.
\newblock In {\em ICCV}. IEEE, 2015.

\bibitem{simonyan2014very}
K.~Simonyan and A.~Zisserman.
\newblock Very deep convolutional networks for large-scale image recognition.
\newblock {\em arXiv preprint arXiv:1409.1556}, 2014.

\bibitem{szegedy2014going}
C.~Szegedy, W.~Liu, Y.~Jia, P.~Sermanet, S.~Reed, D.~Anguelov, D.~Erhan,
  V.~Vanhoucke, and A.~Rabinovich.
\newblock Going deeper with convolutions.
\newblock In {\em CVPR}. IEEE, 2015.

\bibitem{tolias2015particular}
G.~Tolias, R.~Sicre, and H.~J{\'e}gou.
\newblock Particular object retrieval with integral max-pooling of cnn
  activations.
\newblock In {\em ICLR}, 2016.

\bibitem{wah2011caltech}
C.~Wah, S.~Branson, P.~Welinder, P.~Perona, and S.~Belongie.
\newblock The caltech-ucsd birds-200-2011 dataset.
\newblock 2011.

\bibitem{wang2014deep}
X.~Wang, L.~Zhang, L.~Lin, Z.~Liang, and W.~Zuo.
\newblock Deep joint task learning for generic object extraction.
\newblock In {\em NIPS}, pages 523--531, 2014.

\bibitem{wei2016mask}
X.-S. Wei, C.-W. Xie, and J.~Wu.
\newblock Mask-cnn: Localizing parts and selecting descriptors for fine-grained
  image recognition.
\newblock In {\em NIPS}, 2016.

\bibitem{xiao2014application}
T.~Xiao, Y.~Xu, K.~Yang, J.~Zhang, Y.~Peng, and Z.~Zhang.
\newblock The application of two-level attention models in deep convolutional
  neural network for fine-grained image classification.
\newblock In {\em CVPR}. IEEE, 2015.

\bibitem{xie2015fine}
L.~Xie, J.~Wang, B.~Zhang, and Q.~Tian.
\newblock Fine-grained image search.
\newblock {\em TMM}, 17(5):636--647, 2015.

\bibitem{yang2015large}
L.~Yang, P.~Luo, C.~Change~Loy, and X.~Tang.
\newblock A large-scale car dataset for fine-grained categorization and
  verification.
\newblock In {\em CVPR}, pages 3973--3981. IEEE, 2015.

\bibitem{yue2015exploiting}
J.~Yue-Hei~Ng, F.~Yang, and L.~S. Davis.
\newblock Exploiting local features from deep networks for image retrieval.
\newblock In {\em CVPR}, pages 53--61. IEEE, 2015.

\bibitem{zhangspda}
H.~Zhang, T.~Xu, M.~Elhoseiny, X.~Huang, S.~Zhang, A.~Elgammal, and D.~Metaxas.
\newblock Spda-cnn: Unifying semantic part detection and abstraction for
  fine-grained recognition.
\newblock In {\em CVPR}. IEEE, 2016.

\bibitem{zhang2014part}
N.~Zhang, J.~Donahue, R.~Girshick, and T.~Darrell.
\newblock Part-based r-cnns for fine-grained category detection.
\newblock In {\em ECCV}, pages 834--849. Springer, 2014.

\end{thebibliography}

\end{document}